# A real-time analysis of rock fragmentation using UAV technology

Bamford[1] T., Esmaeili[1] K., Schoellig[2] A. P.
[1] Lassonde Institute of Mining, University of Toronto, Toronto, Canada
[2] University of Toronto Institute for Aerospace Studies, Toronto, Canada

*Abstract-* Accurate measurement of blast-induced rock fragmentation is of great importance for many mining operations. The post-blast rock size distribution can significantly influence the efficiency of all the downstream mining and comminution processes. Image analysis methods are one of the most common methods used to measure rock fragment size distribution in mines regardless of criticism for lack of accuracy to measure fine particles and other perceived deficiencies. The current practice of collecting rock fragmentation data for image analysis is highly manual and provides data with low temporal and spatial resolution. Using Unmanned Aerial Vehicles (UAVs) for collecting images of rock fragments can not only improve the quality of the image data but also automate the data collection process. Ultimately, real-time acquisition of high temporal- and spatial-resolution data based on UAV technology will provide a broad range of opportunities for both improving blast design without interrupting the production process and reducing the cost of the human operator.

This paper presents the results of a series of laboratory-scale rock fragment measurements using a quadrotor UAV equipped with a camera. The goal of this work is to highlight the benefits of aerial fragmentation analysis in terms of both prediction accuracy and time effort. A pile of rock fragments with different fragment sizes was placed in a lab that is equipped with a motion capture camera system (i.e., a high-accuracy indoor GPS-like system) for precise UAV localization and control. Such an environment presents optimal conditions for UAV flight and thus, is well-suited for conducting proof-of-concept experiments before testing them in large-scale field experiments. The pile was photographed by a camera attached to the UAV, and the particle size distribution curves were generated in almost real-time. The pile was also manually photographed and the results of the manual method were compared to the UAV method.

## 1. Introduction

Measuring post-blast rock fragmentation is important to many mining operations. Production blasting in mining operations acts to reduce the size of rock blocks so that the rock can be transported from an in-situ location to downstream mining and comminution processes. The rock size distribution induced by blasting influences the efficiency of all downstream mining and comminution processes [1]. It has been shown that rock fragmentation can influence the volumetric and packing properties of the rock (e.g., the fill factor and bulk volume) and, consequently, the efficiency of digging and hauling equipment [2]. Similarly, there have been a number of studies that demonstrate the direct influence of the rock size distribution fed into the crushing and grinding processes on energy consumption, throughput rates and productivity of these processes [1,2]. Due to these impacts, the measurement of post-blast rock fragmentation is an important metric in the optimization of a mining operation. It is suggested that real-time fragmentation measurement should be implemented to improve blast design over time with the goal of producing an optimal rock size distribution for downstream processes [3].

Throughout the history of mining, there have been many methods developed for estimating rock size distribution. The common methods are: visual observation, sieve analysis and image analysis. Visual observation involves inspecting the rock pile and subjectively judging the quality of the blast. This subjective method can lead to inaccurate results. Sieve analysis involves taking a sample of the rock pile being studied and passing it through a series of different size sieve trays. The rock size distribution is calculated by measuring the mass or volume of the rock material that remains on each tray. This method generates more consistent results; however, it is more expensive, time consuming and in certain cases impractical to perform as the sample rock size distribution may not be statistically representative of the whole rock pile. Image analysis methods have been developed with the rise of computer image processing and analysis tools. Conducting image analysis involves taking 2D photos, stereo images or 3D laser scans of the rock pile, and processing these images to determine particle sizes [4-6]. Image analysis techniques enable practical, fast, and relatively accurate measurement of rock fragmentation. However, the following limitations of image analysis have been identified [4]:

- Delineation of particles might be limited due to disintegration and fusion of particles.
- Transformation of surface measurements of particles into volumes may not be representative of the particles being sampled.
- The resolution of the image system is limited compared to that of sieve analysis. Accuracy of the fines regions using image analysis can be very low if the photo captured is not of high enough resolution.
- Mesh sizes assigned to certain rock sizes in image analysis may be different than that assigned in sieving due to the effect of particle shape.
- A constant density is generally applied to all particle sizes so that volume distributions in image analysis are directly related to mass distributions.

In a study of image analysis accuracy, Sanchidrián et al. [4] found that image analysis methods resulted in an error of less than 30% in the coarse region of the rock size distribution. In the same study, an error of less than 85-100% was calculated for the fine region which means that image analysis is not reliable for fine particles. Regardless of these limitations, image analysis is still the most common method used to measure rock fragmentation in mines. The most common image analysis technique applied in mines uses 2D fixed cameras located *(i)* at the base of a rock pile, *(ii)* on shovels and truck buckets, *(iii)* at crusher stations, or on conveyors in the processing plant to capture photos [7-9]. These 2D image analysis techniques have the following limitations:

*(i)* Fixed single camera located at the base of a muck pile:
- Technicians must place scaling objects on the rock pile.
- Photos have to be taken at a distance of less than 20m from the rock pile. This can interrupt production and may place technicians at risk [5].
- The shape of the muck pile can influence the accuracy of the image analysis.
- Only a limited dataset can be collected from a fixed location [8].
- Dust, fog, rain, snow and particulates can obstruct the image.
- Lighting conditions can drastically impact the results of the image analysis [5, 8].

*(ii)* Fixed single camera mounted on shovel booms or truck buckets [8]:
- This requires installing a camera with a clear view at a perspective that is perpendicular to the shovel bucket, which can be difficult.
- Equipment generates large amounts of vibration and shock during operation which can influence the quality of images.
- Shielding is required to protect the camera from falling debris and direct sun light.
- Lighting may not be controlled adequately.
- If truck or shovel is down, no data is collected.
- Imaging the same material multiple times biases the results.

*(iii)* Fixed single camera installed in crusher stations [9]:
- Detailed masking of images is required.
- Scale object must be visible in image.
- Difficult to match material with source.
- Large amount of dust generation obstructs the image.
- Imaging the same perspective multiple times biases the results.

To overcome some of these limitations, 3D measurement techniques have been proposed that use LIDAR stations or stereo cameras to capture images [5, 6, 10, 11]. Using 3D measurements for rock fragmentation analysis eliminates the need for scale objects and reduces the error produced by the shape of the muck pile. If measurements are taken with a LIDAR station, then the error produced by uneven and suboptimal lighting conditions can be eliminated [5] as well. While these techniques reduce the limitations imposed by 2D photos, there are still aspects that can be improved. One example of this is the significant capture time required to take detailed images with a LIDAR system [12]. Another limitation of these 3D imaging techniques is that they are currently limited to capturing images from a fixed location since motion blur can significantly smooth out the 3D data, making particle delineation difficult [10].

In summary, the process of using cameras or LIDARs for post-blast rock fragmentation is highly manual and results in measurements that have low temporal and spatial resolution. Furthermore, there is no current work, to the best of our knowledge, which has focused on determining an optimal image collection procedure for rock fragmentation analysis. To overcome these limitations and to automate the data collection process, this paper presents the use of Unmanned Aerial Vehicle (UAV) technology to conduct real-time rock fragmentation analysis.

In recent years, UAV technology has been introduced into the mining environment to conduct terrain surveying, monitoring and volume calculations [13-16]. These tasks are essential to the mining operation, but they do not leverage all of the benefits that UAVs can offer [15]. UAV technology has the potential to provide acquisition of high resolution data which can be beneficial in blast design, mill operations, and other mine-to-mill process optimization campaigns. In addition, UAVs can provide data acquisition fast and often, which improves the statistical reliability of measurements.

This paper presents the results of a series of proof-of-concept, laboratory-scale tests to measure rock fragmentation using UAVs at the University of Toronto Institute for Aerospace Studies' (UTIAS) indoor robotics lab. The hardware choices, lab configuration, and the procedure used to conduct image analysis are presented. We also discuss the results of the experiments, the benefits of utilizing UAV technology for rock fragmentation measurement, and the image analysis strategy that was developed to achieve optimal image analysis results.

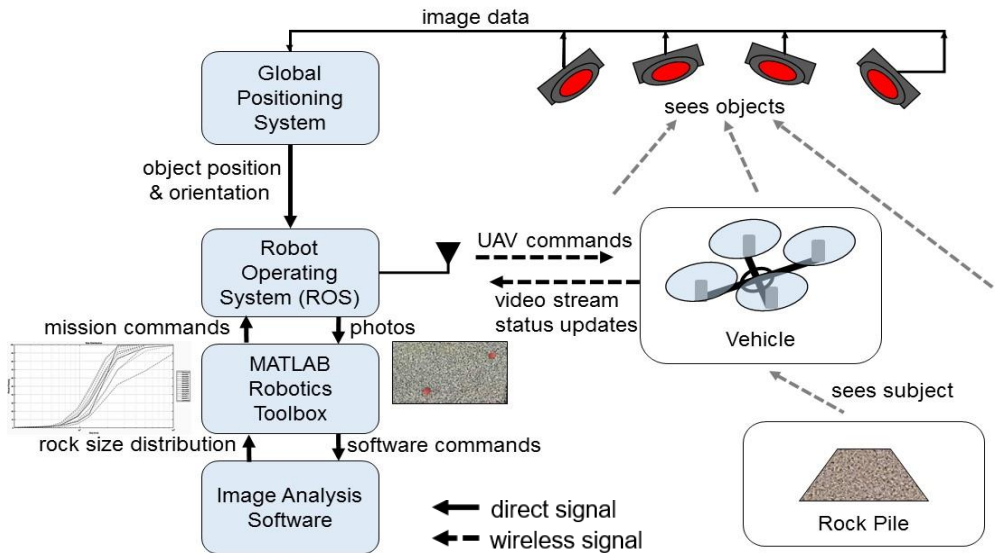

Figure 1: Block diagram of the lab configuration with arrows showing the typical information flow.

## 2. Experiment Setup and Methods

### 2.1. Experiment setup

In order to provide optimal conditions for automated UAV flight for proof-of-concept experiments, demonstrating the feasibility and benefits of automated aerial rock fragmentation analysis, a laboratory experiment was designed and set up. This step was deemed to be necessary before conducting any tests in large-scale field experiments. Figure 1 illustrates the components and overall lab configuration used for the proposed automated rock fragmentation analysis. Figure 2 is a photo taken of the UAV and the lab setup prior to take-off.

#### 2.1.1. Global positioning system

The indoor robotics lab is equipped with a motion capture camera system for precise UAV localization and control. This commercially available system uses ten 4-megapixel Vicon MX-F40 cameras and reflective markers attached to each subject to measure position and orientation at a rate of 200 Hz. For these experiments, the rock pile's and the UAV's position and orientation are collected and sent to the Robot Operating System (ROS) to control the motion of the UAV relative to the pile [17]. Figure 3 shows a screenshot of the Vicon system with the UAV's and rock pile's location plotted. For outdoor practical applications, the camera-based system can be replaced by standard (differential) GPS, a simultaneous localization and mapping (SLAM) solution using onboard cameras for localization [18], or novel alternative localization methods such as the ones based on ultra-wideband [19].

#### 2.1.2. Rock fragment pile

A pile of rock fragments with different sizes, ranging from coarse gravel to fine sand, was built in the lab. Prior to forming the pile, the rock fragments were put through sieve analysis to determine the true rock size distribution as a baseline for the experiments. Locally sourced gravel and sand was collected for sieve analysis. The results of the sieve analysis are presented in Table 1. Once the sieve analysis was completed, the rock fragments were placed on a cart built for use in the indoor robotics lab. Careful attention was given to ensuring that no contamination or material loss occurred during the sample transportation and storage.

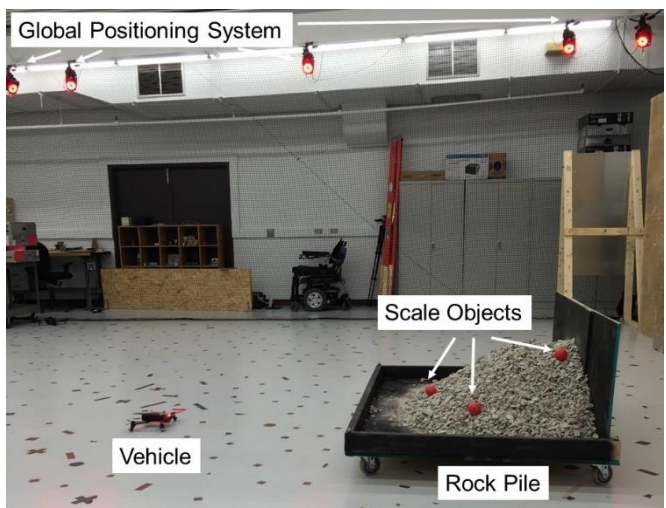

Figure 2: Photo of the lab configuration prior to takeoff.

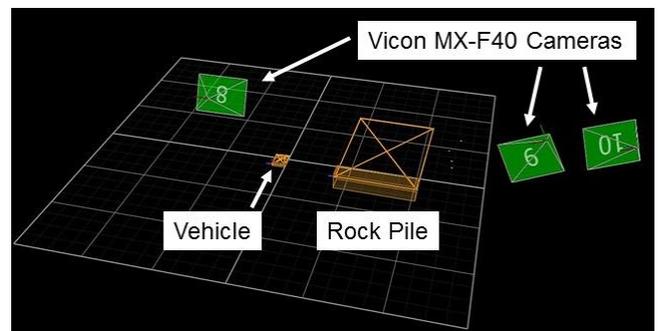

Figure 3: Screenshot of the global sensing system with the UAV and rock pile labeled.

Table 1: Sieve analysis results.

| Mesh Size (mm) | Weight (kg) | % of Total | % Passing |
|---|---|---|---|
| Fines | 1.545 | 0.42% | 0.00% |
| 4.00 | 30.140 | 8.12% | 0.42% |
| 9.53 | 28.535 | 7.69% | 8.54% |
| 12.70 | 167.270 | 45.07% | 16.22% |
| 19.05 | 143.680 | 38.71% | 61.29% |
| Total | 371.170 | 100.00% | |

To use this sieve analysis baseline in the statistical analysis of the manual and automated image analysis methods, a rock size distribution curve was fit to the collected data. The three-parameter Swebrec function [20] was found to be an excellent fit to the data and predicted the coarse region of data much more accurately than the Rosin-Rammler function [21]. The Swebrec function is given by:

$$P(<x) = \frac{1}{1+f(x)} \quad (1) \text{ with}$$

$$f(x) = [ln(x_{max}/x)/ln(x_{max}/x_{50})]^b, \quad (2)$$

where $P(<x)$ is percent passing, $x$ is the rock fragment size, $x_{max}$ is the largest fragment size in the distribution, $x_{50}$ is the size at 50% passing, and $b$ is a curve-modulation factor. To best fit the curve, we find the optimal curve shaping parameters $x_{max}$, $x_{50}$, and $b$. A plot of the sieve analysis results and the Swebrec function fitted to the data is plotted in Figure 4.

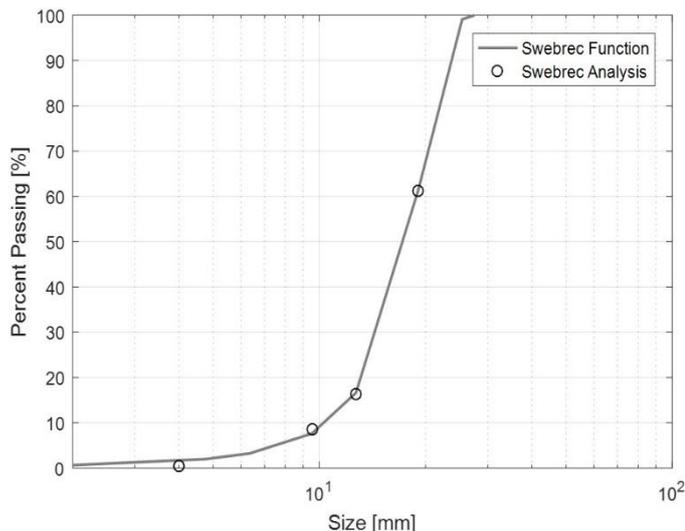

Figure 4: Rock size distribution of sample rock pile plotted with fitted Swebrec curve. Curve fit parameters: $x_{max}$= 27.53 mm, $x_{50}$= 17.84 mm, $b$= 2.79.

### 2.1.3. Drone specifications

A commercially available UAV with integrated camera, the Parrot Bebop 2, was used in our experiments. Table 2 lists the main specifications of the UAV. This UAV has the ability to capture high-resolution photos and videos, which is essential for accurate image analysis. It also has a GPS receiver, which allows us to use it for outdoor field experiments in the future. In this experiment, the UAV broadcasts a secure Wi-Fi network to receive control commands and transmit the video stream to the Robot Operating System (ROS), see Figure 1.

Table 2: Parrot Bebop 2 specifications [22].

| | |
|---|---|
| Camera resolution | 14 megapixels |
| Video resolution | 1920 x 1080 pixels, 30 frames per second |
| Flight time | Approx. 25 minutes |
| Operating range | Depends on Wi-Fi controller device, up to 2 km |
| Battery | Lithium polymer 2700 mAh |
| Flash storage | 8 GB |
| Weight | 500 g |
| Networking | Wi-Fi MIMO Dual Band 2.4 & 5GHz |

### 2.1.4. Lab environment

The indoor robotics lab has fluorescent lighting, which provides optimal lighting conditions for this image analysis experiment. The lab environment is free of wind, which provides optimal conditions for UAV flight. Netting has been installed around the perimeter of a space with dimensions of 10 m x 10 m x 3 m for operator and vehicle safety, see Figure 2. All of these features allow for testing new ideas quickly and safely, and is therefore an ideal lab environment for proof-of-concept experiments.

### 2.1.5. Rock fragmentation image analysis

For these experiments, Split-Desktop, an industry standard software for image analysis in mining, was used [23]. Live images that were captured from the UAV video stream were automatically imported into Split-Desktop and rock fragmentation was computed using appropriate macros and automation scripts. Once the image analysis was completed, rock size distribution information was exported from Split-Desktop to MATLAB for statistical analysis. To determine the size of rock particles, scale objects were required to be placed within the image as a reference. The main software parameters, such as the fines factor, were calibrated using sieve analysis data. The fines factor, used for each image, was zero and the scale object size was set to 60 mm.

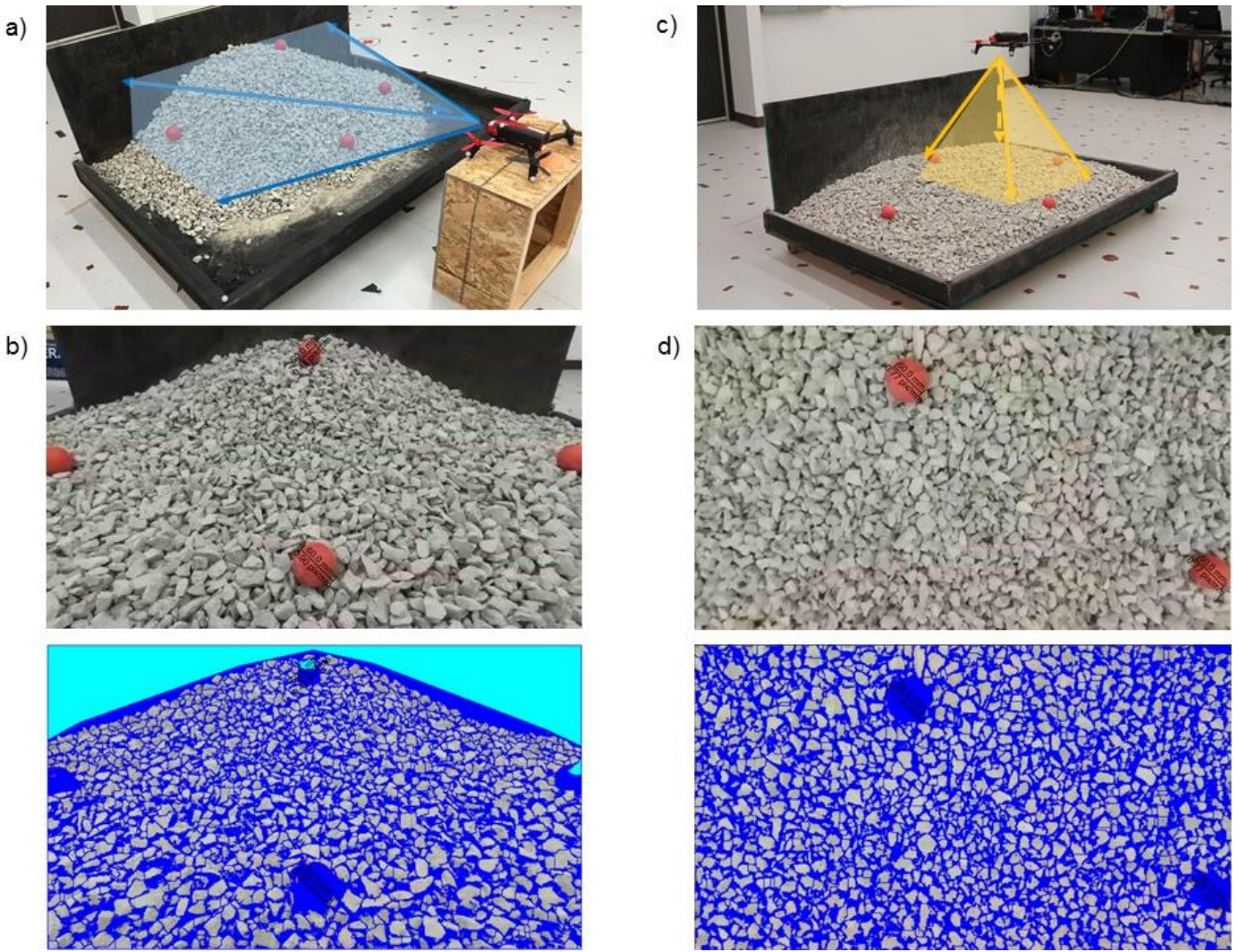

Figure 5: a) UAV set up as a fixed camera for manual image analysis. b) Raw and delineated photo captured in manual image analysis. c) UAV in flight for automated image analysis. d) Raw and delineated photo captured in automated image analysis.

*2.1.6. Robot Operating System (ROS)*

The open-source Robot Operating System (ROS) was chosen to act as the central software node of the experimental setup. ROS is a flexible software framework for writing robot software that has been widely adopted [24]. In these experiments, ROS uses high-level path plan and actual position and orientation measurements from the global positioning system to send low-level velocity and orientation commands wirelessly to the UAV. ROS itself receives sensor data from the UAV and broadcasts it to the network for the subsequent image analysis, see Figure 1.

*2.1.7. MATLAB® Robotics System Toolbox™*

The MATLAB Robotics System Toolbox acts as an interface between ROS and Split-Desktop while providing statistical analysis to the operator in real-time. The Robotics System Toolbox was used to capture and save broadcasted images, call a macro to run image analysis on Split-Desktop, and import the rock size distribution generated by Split-Desktop for statistical analysis.

**2.2. Aerial rock fragmentation analysis with a UAV**

To highlight the benefits of aerial fragmentation analysis in terms of both prediction accuracy and time effort, the automated UAV image analysis was tested in the lab. In addition, an image analysis approach based on a fixed camera – as is typically done in practice – was also tested. This allowed for a direct comparison between these two methods.

To ensure that camera lens bias was not added to the samples, the UAV camera was used for both methods using the same image resolution. In order for images being captured from the same sample surface, a typical rock pile configuration was fixed for both experiments. For comparison, each method's steps were timed, starting at setup and ending at the export of a final rock size distribution. Once these analyses were conducted in the lab, statistical analysis was done to compare each method's predictive accuracy. This comparison was then used to propose an optimal strategy for image analysis of rock fragmentation.

The following subsections describe the procedure that was followed by the operator for the manual and automated image analysis method.

*2.2.1. Fixed-camera, manual image analysis*

When creating this procedure, it was noted that there is no literature that describes an optimal or standard procedure to use while manually capturing images for rock fragmentation image analysis. The procedure used in this work was as follows:
1. place scale objects and prepare UAV camera in front of rock pile (Figure 5a);
2. take photos of the muck from different positions around the rock pile base looking horizontally with ~50% overlap to simulate the current practice used for capturing images at the base of a rock pile (Figure 6);
3. return the UAV camera to the workstation;
4. transfer images to the workstation and remove images that are of poor quality;
5. conduct image analysis using Split-Desktop to obtain rock size distribution (Figure 5b).

*2.2.2. UAV automated image analysis*

For the automated analysis, the procedure was as follows (cf. Figure 1):
1. place scale objects and prepare and initiate automated UAV fragmentation analysis system;
2. if systems are ready and conditions are safe to fly, send command to takeoff;
3. as UAV automatically moves along the predefined path taking two levels of photos with ~50% overlap, ensure that UAV operates safely and intervene if problems occur (Figure 6 and Figure 5c);
4. once the UAV returns to the take-off location, analysis is finished, send command to land;
5. on the MATLAB window, save rock size distribution results after filtering out poor quality images (Figure 5d).

## 3. Results and Discussion

At the time of this paper, multiple trials have been conducted to develop the UAV-based, automated rock fragmentation analysis and to compare it with the conventional, manual method. This paper presents the results of one representative trial. Benefits of using aerial fragmentation analysis are summarized and quantified in Section 3.3. Finally, an optimal strategy of measuring rock fragmentation using UAVs is proposed in Section 3.4.

### 3.1. Summary of collected data

A summary of a typical manual and automated fragmentation analysis experiment are given in Table 3. Eleven photos were taken in the manual, fixed-camera experiment such that an overlap of 50% was achieved between adjacent images. Sixteen photos were taken by the automated UAV method to achieve the same amount of overlap and to capture a small (closer) and medium (farther) scale measurement through holding two different altitudes above the pile (see Figure 6).

Table 3 also includes a list of time entries for each method. These time entries represent the amount of time taken for each step in the procedure described in Section 2.2.1 and 2.2.2 for manual and automated image analysis, respectively. "Preparation" time is the time taken to complete step 1 for manual fragmentation analysis and step 1 for automated analysis. Step 2 in the manual analysis and steps 2-4 in the automated procedure are measured as the "operating" time.

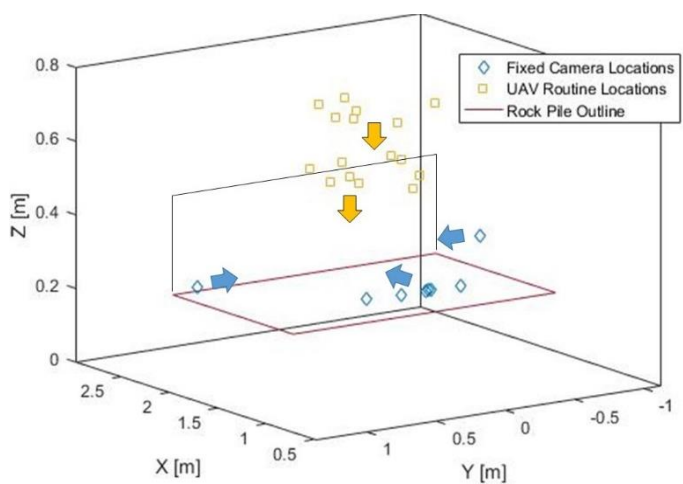
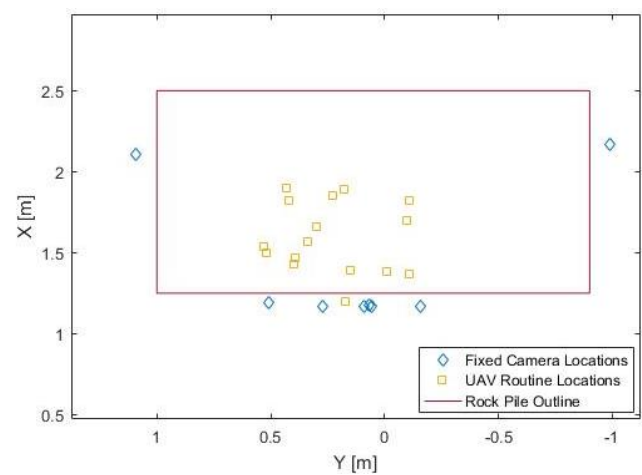

Figure 6: Location and camera direction used to capture images for fixed-camera, manual image analysis (blue) and UAV-based, automated image analysis (yellow).

"Breakdown" is described in steps 3-4 for the manual analysis and step 5 for the automated analysis. "Analysis and editing" time is unique to the manual, fixed-camera method since the conventional technique requires a technician to process the images and analyze results after data is collected (step 5 in Section 2.2.1), where this step is fully automated in the UAV-based procedure. Figure 7a and 7b provide the rock size distribution calculated by the manual, fixed-camera method and the automated, UAV-based method, respectively.

Table 3: Trial information for image analysis methods.

|  | Manual, fixed camera | Automated UAV |
|---|---|---|
| Number of photos taken: | 11 | 16 |
| Number of photos used in analysis: | 10 | 14 |
| **Time Entries** | | |
| Preparation: | 4:13 min | 1:35 min |
| Operating: | 4:19 min | 6:04 min |
| Breakdown: | 3:46 min | 2:23 min |
| Analysis and editing: | 43:34 min | 0:00 min |
| **Total time:** | **55:52 min** | **10:02 min** |

### 3.2. Comparison of manual and automated method

To directly compare the proposed UAV automated image analysis with that of the conventional fixed-camera method, two metrics were considered: time effort and prediction accuracy.

*3.2.1. Time effort*

The total time effort that was expended for each method is given in Table 3. As can be seen, the UAV image analysis method took approximately 20% of the time that the conventional method takes. The fixed-camera method requires a lot of time spent processing images prior to gaining results and after data is collected. The majority of this time is spent preparing images and editing the delineations of particles to reduce fusion and disintegration error. Counter to this, the automated UAV method generates results in real-time during flight. The time difference between methods is expected to be even more pronounced in a field experiment where more data is collected, and consequently, more processing time is taken by the conventional method. The UAV image analysis method for this experiment has not been optimized yet and with further development will reduce operating time by choosing strategic locations to capture statistically significant measurements. For example, we may get optimal results by flying the UAV to only two locations and taking two photos at each location with different camera angles. This type of improvement contrasts the numerous locations that were chosen in this experiment (see Figure 6).

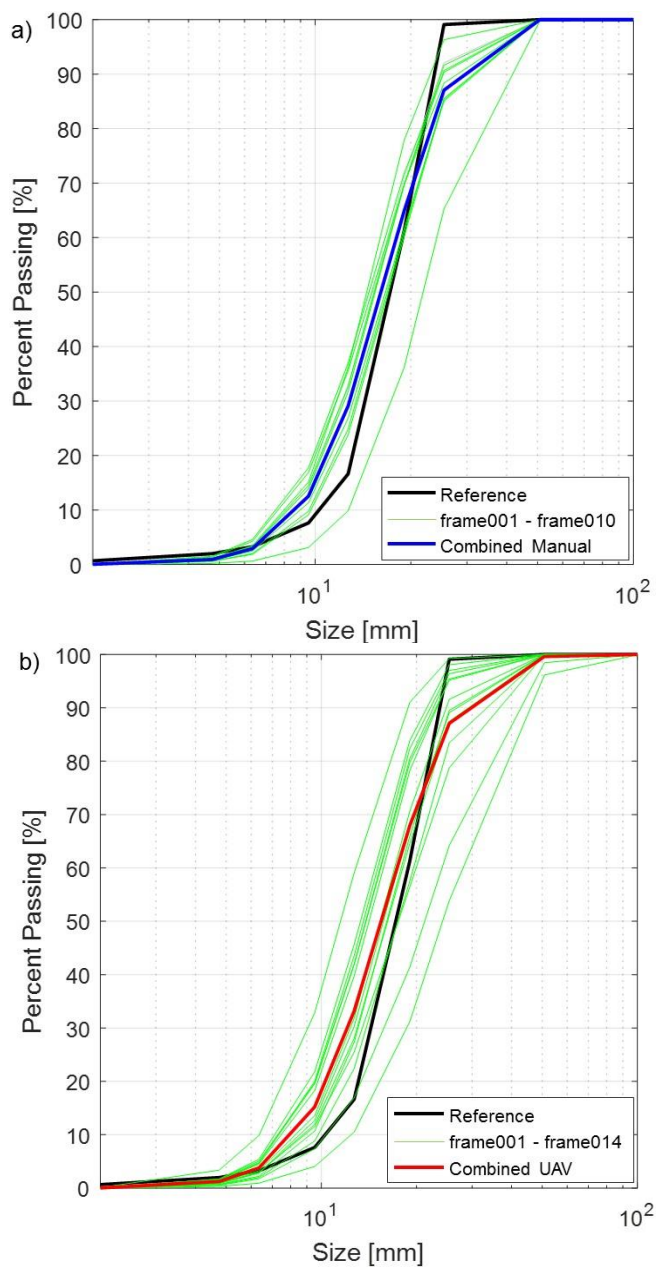

Figure 7: a) Manual, fixed-camera rock size distribution. b) Automated UAV rock size distribution.

*3.2.2. Prediction accuracy*

To determine the prediction accuracy of each method for comparison, the true error in percent passing and characteristic rock size was determined. The percent error of percent passing ($P(<x)$) for each sieve size is given by

$$Percent\ True\ Error = \frac{P(<x)_{Image\ Analysis} - P(<x)_{Sieve\ Analysis}}{P(<x)_{Sieve\ Analysis}} \times 100\%. \quad (3)$$

The resulting error distributions for each method are presented in Figure 8, with the error's standard deviation plotted as bars and the average error plotted as a solid black

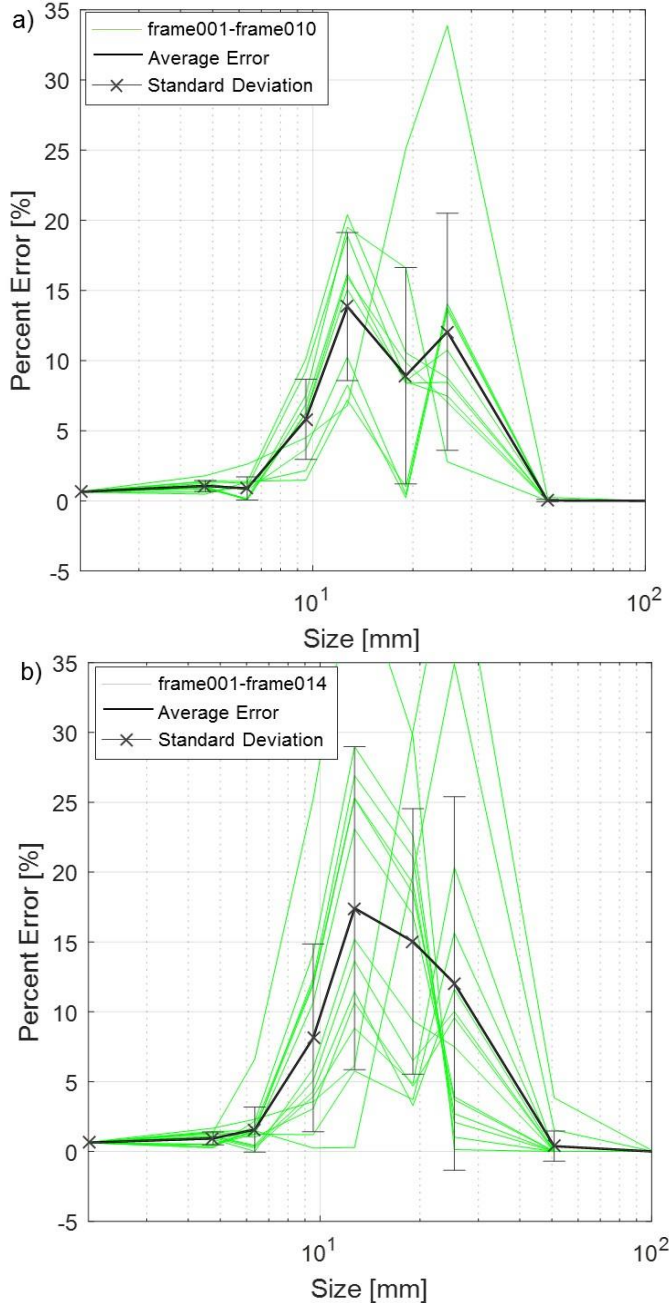

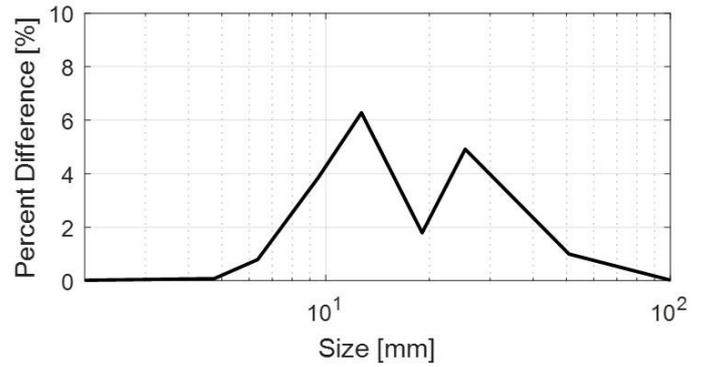

Figure 9: Percent difference between the manual and UAV methods.

orientation and/or minor (possibly automated) editing, this source of error can be eliminated.

The percent difference between these two methods is given by

$$Percent\ Difference = \frac{|P(<x)_{Manual\ Method} - P(<x)_{Automated\ Method}|}{P(<x)_{Manual\ Method}} \times 100\%, \quad (4)$$

which results in a difference ranging between 1-6% over the 2 mm to 1905 mm size range. The largest percent difference occurs at the mid-range size of 12.70 mm due to the narrow distribution of particle sizes in this region (see Figure 9).

Characteristic rock fragment sizes, such as P80, is the rock fragment size for which a percentage of the weight (i.e., 80% for P80) is smaller than. We have chosen three standard characteristic rock fragment sizes to compare the image analysis methods: P80, P50 and P20. To compare the error between these sizes, the percent logarithmic error and average percent logarithmic error were used. For example, the equations for the percent logarithmic error and average percent logarithmic error for P80 are:

$$Percent\ True\ Logarithmic\ Error\ of\ Frame\ i = \frac{\log(P80_{Image\ Analysis}) - \log(P80_{Sieve\ Analysis})}{\log(P80_{Sieve\ Analysis})} \times 100\% \quad (5)\ and$$

$$Average\ Percent\ True\ Logarithmic\ Error = \frac{\sum_{i=1}^{n} Percent\ True\ Logarithmic\ Error\ of\ Frame\ i}{n}, \quad (6)$$

where $n$ is the number of frames taken since the start of the analysis. To find these values, the characteristic sizes for each photo had to be determined. This was done by fitting a Swebrec function to the rock size distribution estimated by the image analysis method and rearranging Equation 1 to solve for the characteristic size. Figure 10 gives a series of plots which illustrate the average characteristic size error and the characteristic size error by photo frame. Again, the automated UAV image analysis method has a larger variation of error. However, the average characteristic size error of the automated line. For these plots, it is apparent that the UAV method has more variation in results. Through interpretation of the images collected, this variation may result from suboptimal scale placement and the photo's location and camera angle. In turn, this may mean that the location and orientation of pictures taken by a UAV should be further optimized. In addition to this, some of the images captured by the UAV contained the rock pile edges and floor. In the aerial method, these were treated as large particles and contributed to the variance of the UAV method. This is an error introduced by the experiment set-up. With an optimized combination of picture location and method is within 2-5% of the conventional method. In some cases, such as for P80, the automated method gave a better prediction than the fixed-camera method. It is also interesting that with more photos taken in the UAV method, the error reduces, whereas the error for the manual method stays at about the same level. Intuitively, this could mean that gathering more images in the automated UAV method may reach an error that is less than that of the conventional method. By expanding this intuition, it may be possible to determine a minimum number of photos that would be required to reach a desired threshold of error. These ideas will be further analyzed in the future.

Figure 8: a) True error distribution of the fixed-camera method. b) True error distribution of the automated UAV method.

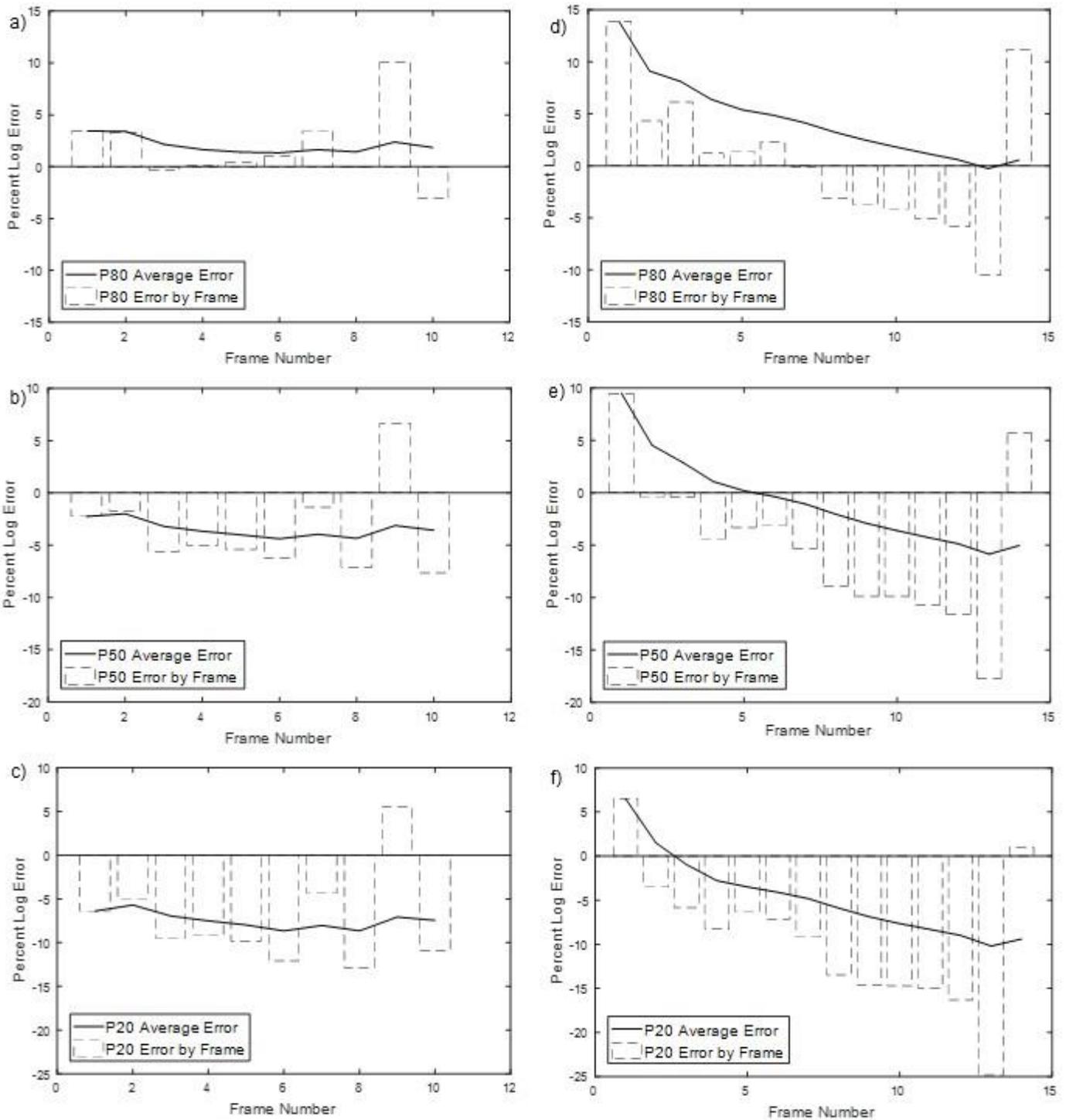

Figure 10: a-c) Characteristic size error and average error using the first *n* frames calculated for the fixed-camera image analysis method. d-f) Characteristic size error and average error using the first *n* frames calculated for the automated UAV image analysis method. From top to bottom: error in P80, P50, and P20.

Overall, based on our preliminary analysis, the automated UAV analysis method performed better than the conventional method in terms of time effort (five times faster) and, on average, predicted the rock size distribution within 17% of the sieving analysis measurement (see Figure 8b). The largest error occurred in the coarse region of the rock size distribution. This automated method also resulted in a size distribution prediction that was within 6% of the manual image analysis method (see Figure 9). This is considered to be very accurate for rock fragmentation image analysis, especially since the findings of [4] suggest that image analysis relative to sieve analysis can reach 30% error in coarse regions and up to or beyond 100% error in the fines region. As a result, the proposed automated, UAV-based technique can provide at least comparable accuracy to the manual methods.

The largest errors produced in this experiment were found to be caused by the scale of the experiment since bin edges interfered with rock size measurement producing error in the coarse fraction. This effect is illustrated in Figure 11. With an optimized combination of picture location and orientation or minor editing of images, this source of error can be eliminated resulting in even higher accuracy of the analysis.

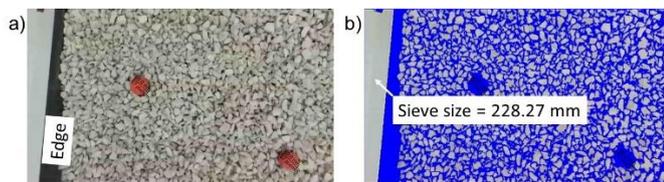

Figure 11: Example of bin edges interfering with rock size measurement in flight: a) before delineation, and b) after delineation, which predicts a rock size of 228mm on the left side.

### 3.3. Discussion of benefits

Throughout the development of the automated aerial fragmentation analysis system, a number of benefits have been identified. The main benefit is that the UAV system collects and analyzes images rapidly. This serves to reduce the cost to the operator and enables on-demand, real-time, high-resolution data collection. On top of this, the system provides results that are considerably accurate. For these reasons, the UAV system is considered a valuable tool for rock fragmentation real-time monitoring strategies.

Current benefits provided by the UAV system are:
- Collection of data does not interrupt the production process.
- UAV is capable of sampling regions of interest that are otherwise inaccessible by a human operator.
- Results are available in real-time allowing the real-time adjustment of the UAV's flight path to optimize the results of the fragmentation analysis.
- Real-time results also allow the immediate adjustment and optimization of blast designs.
- Surface sampling errors are reduced with high-frequency measurements (e.g., a UAV measurement campaign every eight hours).
- Fragmentation analysis resolution can be easily adjusted to target different regions in the rock size distribution by flying closer or further away from the rock pile.
- Obstruction of the image by particulates can be controlled and avoided.
- Additional data, such as photogrammetry for volume calculations, can be collected simultaneously as part of the UAV mission.
- Sampling bias (resulting from taking the same image multiple times) can be controlled and extreme outliers can be filtered out in real-time.
- The system keeps operator out of harm's way in an active mining environment. A UAV is expendable; the human operator is not.

### 3.4. Possible Future Extensions

Possible extensions of the UAV automated image analysis method that will be investigated in future works include:
- Statistics may be used to determine the number of samples required to reach a desired level of significance (at 5% significance level), and the UAV mission plan can be adjusted accordingly. Preliminary results given in Figure 12 show the required number of images over time using the statistical student's t-test for the characteristic sizes of P80, P50 and P20 [25]. To understand what this plot represents, consider that the technician is most interested in the P80 rock size, at frame 10 the required number of photos for a statistically significant measurement is 11 and at frame 11 the required number of images is 11. Therefore, at frame 11 the UAV mission can be stopped. The coarse region needs more samples to be statistically significant, if the technician were interested in the P20 or P50, the UAV mission could have been stopped earlier with less photos. This method has many limitations but in a practical situation, where sieve analysis is not available, it may help determine the number of photos that would be required to obtain a representative measure of rock fragmentation.
- The shape of the muck pile can be accounted for and the camera angle adjusted accordingly using a combination of laser rangefinders or sonar sensors.
- The flight path and camera angles can be further optimized to achieve better performance on both measures: time effort and prediction accuracy.
- Scale objects can be eliminated through the use of laser rangefinders, onboard stereo cameras, modern sensing devices such as the Intel® RealSense™ Technology, or UAV teams.

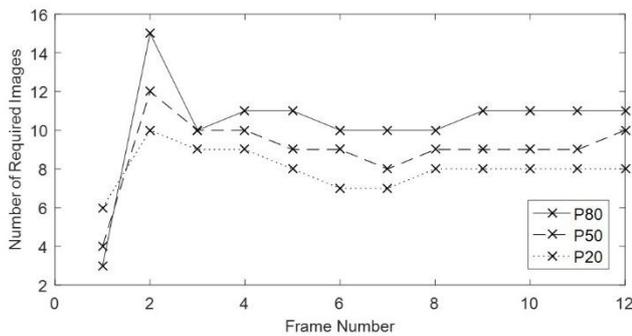

Figure 12: t-test calculating the number of required images over time using an alternate hypothesis that the new mean P80, P50, P20 will be 20% greater than the current mean at 80% power and 5% significance level.

## 4. Conclusion

This paper presented the results of a series of proof-of-concept, laboratory-scale tests to measure rock fragmentation using UAVs. The configuration of an automated UAV system that collects rock fragmentation data in real-time has been described in detail. Procedures for collecting data with the UAV system have been outlined from the perspective of the technician collecting the data. The automated method of collecting rock size distribution information was compared with conventional techniques. UAV technology was shown to only take a fraction of the time (~20%) that a conventional method takes to measure rock fragmentation within 6% of the conventional method's accuracy, where the conventional method deviates from the true distribution by up to 14%. In addition to providing god accuracy results, a number of benefits were identified throughout the study. The main benefit being that UAVs can provide data acquisition fast and often, which improves the statistical reliability of measurements and reduces sampling error, while not interrupting production processes.

Future work will focus on implementing this system in an active mining environment to gain more insight into the system's prediction accuracy, the value added, and its ability to be incorporated into mine-to-mill optimization. At the time of this paper, the authors are investigating the impact of both the UAV's location and camera angle, and the artificial lighting from the UAV on the prediction results. The authors are also investigating the impact of high-frequency measurement during rock pile extraction and its effect on sampling bias. These results will be reported on in a future paper.

**Acknowledgement**

The authors would like to thank Split-Engineering for their generous support of this project. Furthermore, the authors wish to thank the University of Toronto's Dean's Strategic Fund "Centre for Aerial Robotics Research and Education (CARRE)", the Canada Foundation for Innovation John R. Evans Leaders Fund and the Natural Sciences and Engineering Research Council of Canada for their financial support of this project.